\documentclass{article}

\usepackage{arxiv}

\usepackage[utf8]{inputenc} 
\usepackage[T1]{fontenc}    
\usepackage[colorlinks=true, allcolors=blue]{hyperref}
\usepackage{url}            
\usepackage{booktabs}       
\usepackage{amsfonts}       
\usepackage{nicefrac}       
\usepackage{microtype}      
\usepackage{lipsum}
\usepackage{graphicx}
\graphicspath{ {./images/} }
\usepackage{authblk}
\usepackage{amsmath,amssymb, amsthm, mathtools, bbm}
\usepackage{multirow}
\usepackage{rotating}
\usepackage{caption}
\usepackage{comment}
\usepackage{xcolor}

\theoremstyle{definition}
\newtheorem{definition}{Definition}[section]

\title{A Novel Patch-Based TDA Approach for Computed Tomography Imaging}

\author[a]{Dashti A. Ali}
\author[b]{Aras T. Asaad}
\author[c]{Jacob J. Peoples}

\author[d]{Ahmad Bashir Barekzai}
\author[c]{Camila Vilela}
\author[c]{Hala Khasawneh}

\author[c]{Jayasree Chakraborty}
\author[c]{João Miranda}
\author[a]{Mohammad Hamghalam}
\author[c]{Natalie Gangai}
\author[c]{Natally Horvat}

\author[c]{Richard K. G. Do}
\author[d]{Alice C. Wei}
\author[e,f]{Amber L. Simpson}

\affil[a]{School of Computing, Queen’s University, Kingston, ON, Canada}
\affil[b]{School of Computing, The University Of Buckingham, Buckingham, United Kingdom}
\affil[c]{Department of Radiology, Memorial Sloan Kettering Cancer Center, New York, USA}
\affil[d]{Hepatopancreatobiliary Service, Department of Surgery, Memorial Sloan Kettering Cancer Center, New York, USA}
\affil[e]{Department of Radiology and Diagnostic Imaging, University of Alberta, Edmonton, AB, Canada}
\affil[f]{Alberta Machine Intelligence Institute, Edmonton, AB, Canada}

\begin{document}
\maketitle

\begin{abstract}
The development of machine learning models based on computed tomography (CT) imaging has been a major focus due to the promise that imaging holds for diagnosis, staging, and prognostication. These models often rely on the extraction of hand-crafted features where incorporating robust feature engineering improves the performance of these models. Topological data analysis (TDA), based on the mathematical field of algebraic topology, focuses on data from a topological perspective, extracting deeper insight and higher dimensional structures. Persistent homology (PH), a fundamental tool in TDA, extracts topological features such as connected components, cycles, and voids. A popular approach to construct PH from 3D CT images is to utilize 3D cubical complex filtration, a method adapted for grid-structured data. However, this approach is subject to poor performance and high computational cost with higher resolution images. This study introduces a novel patch-based PH construction approach designed for volumetric CT imaging data that improves performance and reduces computational time. This study conducts a series of experiments to comprehensively analyze the performance of the proposed method and benchmarks against the cubical complex algorithm and radiomic features. Our results highlight the dominance of the patch-based TDA approach in terms of both classification performance and computational time. The proposed approach outperformed the cubical complex method and radiomic features, achieving average improvement of 7.2\%, 3.6\%, 2.7\%, 8.0\%, and 7.2\% in accuracy, AUC, sensitivity, specificity, and F1 score, respectively, across all datasets. Finally, we provide a convenient Python package, Patch-TDA, to facilitate the utilization of the proposed approach.
\end{abstract}

 
\section{Introduction}
The development of computer aided diagnostic tools has gained significant interest in diagnostic imaging as they offer the potential to better inform clinical decision making from standard-of-care medical images. Computed tomography (CT) scans are, in particular, a primary source of data to build robust machine learning (ML) models because they provide high resolution volumetric information. Consequently, many ML methods have been proposed that rely on hand-crafted radiomic features or sophisticated deep learning-based approaches that have shown success in numerous clinical tasks. However, deep learning methods are difficult to explain due to black-box decision-making and typically require high computational demands, such as GPU acceleration \cite{hayashi2019right, vial2018role, majeed2020issues}. Radiomic features, although not GPU-dependent, rely on pixel-wise comparisons, and are therefore sensitive to variations in image acquisition settings, such as differences in image resolution or contrast uptake~\cite{mayerhoefer2020introduction, lee2021radiomics, rizzo2018radiomics}. Topological data analysis (TDA) has emerged as a promising approach to address some of these limitations, providing a unique way to extract richer patterns and insight from data. TDA is built mainly on theoretical concepts from a field of mathematics called algebraic topology, analyzing the topological structure of data across different scales, revealing complex relationships and uncovering hidden patterns that may not be visible to the naked eye \cite{carlsson2009topology, lum2013extracting}.

Robust feature engineering approaches are vital to the development of ML models for volumetric imaging to take advantage of the temporal information present in the data. Cubical complex filtration is a standard TDA approach to extract topological features by constructing persistent homology (PH)~\cite{somasundaram2021persistent, tanabe2021homological}. Although this approach has shown success on 2D images, further improvements are needed to efficiently capture 3D topological features in volumetric images. In this study, we propose a patch-based PH construction method to address these challenges. This approach focuses on the transformation of the volumetric data to an $d$-dimensional point cloud through the process of 3D image patch summarization to a $d$-dimensional point. With a point cloud representing the 3D image, we have utilized alpha complex filtration, an efficient approach to construct PH from point cloud data. Note that another advantage of this approach is that, unlike cubical complex, with point cloud data higher dimensional structures beyond connected components, loops and enclosed voids can be explored. In our previous work \cite{ali2026a}, a PCA-based approach was proposed which only utilized PCA for patch intensity summarization. The current study provides different patch intensity summarization approaches and also integrates compressed patch coordinate representation, which outperforms our previous approach.

TDA has demonstrated significant success in the development of models based on volumetric medical imaging modalities such CT and MRI \cite{singh2023topological}. For example, in the assessment of tumour heterogeneity in lung adenocarcinomas based on CT images \cite{kawata2021representation}, analysis of osteoarthritis from MRI images \cite{pedoia2018mri}, classification of chronic obstructive pulmonary disease and explaining the geometric structure of the airway system through lung CT images \cite{belchi2018lung}, evaluation of the accuracy of hepatic cancer prediction via MRI images  \cite{oyama2019hepatic}, distinguishing squamous cell carcinoma from adenocarcinoma and benign from malignant tumours in lung CT images \cite{vandaele2023topological}, unveiling the dynamical organization of the brain using fMRI data \cite{saggar2018towards} and characterization of the 3D structure of trabecular bone and predicting bone strength using bone MR images \cite{gomberg2000topological}. To the best of our knowledge, the proposed technique is the first approach to outperform classical cubical complex–based method for PH computation
on volumetric medical imaging data by more effectively exploiting three-dimensional information.

The first contribution of this paper is the patch-based PH construction approach for 3D data that achieves both better classification performance and efficiency than the classical cubical complex filtration approach. The second contribution is the comprehensive analysis of different patch-to-point transformation techniques to transform volumetric data to a point cloud. This includes an extensive set of experiments on four CT datasets for different clinical classification tasks, such as predicting patient response to therapy. The third contribution of this study involves a systematic benchmarking of the proposed patch-based PH construction approach against two baselines---the classical 3D cubical complex filtration method and radiomic features---with respect to classification performance. This contribution further includes benchmarking the proposed method against the cubical complex-based approach in terms of PH computation efficiency. Finally, the implementation of the proposed patch-based TDA algorithm is encoded into a convenient python package to facilitate user experimentation with the approach.

\section{Methods}
\subsection{Topological Data Analysis}
Topology is a field of mathematics concerned with the connectivity and continuity properties of objects. An important aspect of the topological spaces and their properties is the ability to be coordinate free, invariant under continuous deformation, and to be expressed in compressed form \cite{lum2013extracting}. Using tools in the field of TDA, we can extract topological invariants of complex and high dimensional data which are stable in terms of small perturbations to input noise and can be employed with various ML tasks such as classification. Topological invariant examples studied in this work are connected components, loops, and cavities (i.e., enclosed voids).

These topological invariants can be quantified by Betti numbers. The first $N$ Betti numbers are used to differentiate objects with differing topology in $N$-dimensional space. For instance, the Betti number of 2D objects includes $(\beta_0, \beta_1)$, and $(\beta_0, \beta_1, \beta_2)$ for 3D, where $\beta_0$, $\beta_1$ and $\beta_2$ refer to connected components, loops and voids respectively. Although this characterization is simple to describe in abstraction, the practical computation requires a smart machinery, which can be achieved through homology theory, a well-established technique in algebraic topology. To apply homology theory to compute Betti numbers of objects, they must be represented in combinatorial structures such as simplicial or cubical complexes \cite{otter2017roadmap}.

\begin{figure}[h]
\begin{center}
\includegraphics[width=0.8\linewidth]{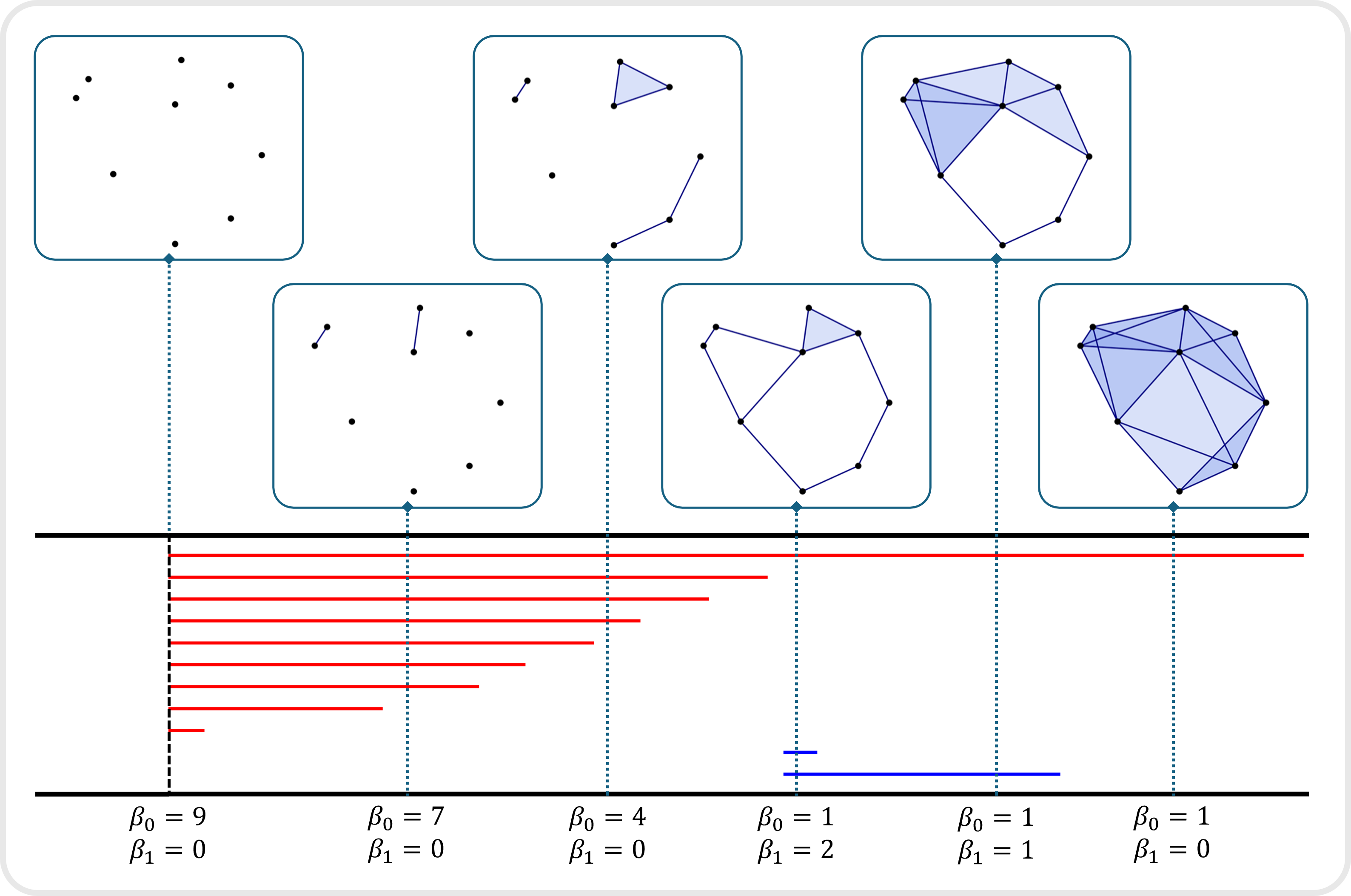}
\end{center}
\caption
{ \label{fig:simplicial_complex} 
Generating persistent barcodes from a sample point cloud. In the far left side, at the beginning, the distance threshold is zero, hence each point is an independent component (nine red bars start at zero correspond to nine points). Each red bar tracks the evolution of a homological feature in dimension zero (connected components). As we move to the right, the threshold is increased and points, whose distance are within the threshold, will connect (merge into one component). Eventually, all the points are merged into a single connected component. During this process, loop structures, homological feature in dimension one, are also forming and eventually vanishing as the threshold is increased (indicated by blue bars). The start and end of a bar indicate the birth and death of a feature, respectively, and the longer the bar, the more persistent the feature is. The collection of the red bars is referred to as PBs in dimension zero, while the blue bars forms PBs in dimension one.}
\end{figure}

PH is a principal tool in the field of TDA, used to quantify the shape of data at different scales (computing Betti numbers), tracking the evolution of topological features such as connected components, cycles and cavities. Homology provides the theoretical framework for persistent homology. In order to compute topological features from data, one need to utilize a combinatorial approach, known as filtration, such as a simplicial or cubical complex, to build PH. A simplicial complex (SC) is a topological space build from a set of multi-dimensional triangular subsets of $\mathbb{R}^n$, by joining some of them together along their boundary components such as nodes, edges and faces. SCs transform the continuous domain of topological space to the discrete domain of data \cite{nanda2013simplicial}. Figure \ref{fig:simplicial_complex} demonstrate this process on a sample point cloud. The evolution of topological structures, including connected components and loops can be observed from the figure as the filtration process advances from left to right,  by increasing the distance threshold. Through this process, each topological feature is tracked using a horizontal bar, where the start of the bar indicates when the feature first appears, and the end of the bar indicates when the feature disappears or merges with another feature. A collection of such bars is called a persistence barcode (PB). The concept of PBs will be explained further in later sections. In general, the process of acquiring topological features from data involves two main steps. First constructing PH from the data through a filtration method to obtain PBs, and second, featurizing the space of PBs using vectorization methods to obtain feature representations suitable for training ML models. In this work, we utilize two distinct filtration approaches: alpha complex for building PH from point cloud data and the cubical complex, which is tailored for grid-structured data such as 2D digital images. In the following sections, these methods will be explained in detail.

\subsection{Alpha Complex}
In the context of PH, alpha complex provides a computationally efficient approach to build multi-resolution topological features of a set of $n$-dimensional points. Alpha complex is a family of sub-complexes from the Delaunay triangulations of a point set and provides a geometrically meaningful approximation of the union of balls centred at the input points. As the parameter $\varepsilon$ increases, the radius of these balls grows, and new simplices are added to the complex in a manner that respects the geometry of the data. Next, we describe alpha complex mathematically together with necessary definitions. 

\begin{definition}[Cover]
Let $X=\bigcup_{i\in I}U_i$ be a topological space. Then, the cover of $X$ is defined as a non-empty and finite collection of open sets $\mathcal{U}=\{U_i\}_{i\in I}$. More specifically, $\mathcal{U}$ is called a \textit{good cover} if all individual sets $U_i \in \mathcal{U}$ and all finite, non-empty intersections of those sets are contractible, i.e., they can be continuously deformed into a single point \cite{carlsson2014topological, bott2013differential}.
\end{definition}

\begin{definition}[Nerve]
The nerve $\mathcal{N}$ of $\mathcal{U}$ can be defined by the following criteria:
\begin{enumerate}
    \item $\phi \in \mathcal{N}$.
    \item If $\bigcap_{j\in J}U_j\neq \phi$ for $J\subseteq I$, then $J\in \mathcal{N}$.
\end{enumerate}
\end{definition}

\begin{definition}[Voronoi cell]
Let $X$ be a finite set of points in $\mathbb{R}^{d}$. The Voronoi cell of a point $x$ in $X$ is the set of points $V_{x}\subseteq\mathbb{R}^{d}$ for which $x$ is the closest point in $X$, i.e. $V_{x}=\{u\in\mathbb{R}^{d}:|u-x| \leq |u-x'|, \forall x' \in X\}$. 
\end{definition}

\begin{definition}[Delaunay complex]
The Delaunay complex of a finite set $X \in \mathbb{R}^{d}$ is the nerve of the corresponding Voronoi diagram.
\end{definition}

The intersection of the Voronoi cell $V_x$ with $B(X, \varepsilon)$ is defined as $\mathbb{R}_\varepsilon(x)$, i.e., $\mathbb{R}_\varepsilon(x) = V_x \bigcap B(X,\varepsilon)$ for every $x\in X$. With these definitions in place, we can now formally define alpha complex.

\begin{definition}[Alpha complex]
The alpha complex $\mathcal{A}_\varepsilon(X)$ is the nerve of the covers shaped by $\mathbb{R}_\varepsilon(x)$ for all $x \in X$, \cite{pun2022persistent} i.e. 
\[
\mathcal{A}_\varepsilon(X) := \{\sigma \in X \mid \bigcap_{x \in \sigma} \mathbb{R}_\varepsilon(x) \neq \phi\}.
\]
\end{definition}

\subsection{Cubical Complex}
Cubical complex is a combinatorial approach to construct PH from grid-structured data, for example, 2D or 3D digital images. Cubical complex is constructed by ordering the cells (vertices, edges, squares, etc.) using a scalar function, voxel intensity value in the context of 3D CT scan images where cells are added to the complex based on the ascending order of their voxel intensity values. Mathematically, a finite cubical complex in $\mathbb{R}^{3}$ is a collection of cubes aligned on the grid $G^{3}$, satisfying some conditions analogous to a SC \cite{garin2019topological}. A $3$-dimensional image can be represented as a map $\eta:I \subseteq G^{3} \rightarrow \mathbb{R}$. For an element $v\in I$, called a voxel (pixel when \(d=2\)), its intensity or grayscale value is given by $\eta(v)$. There are several ways to represent digital images as cubical complexes. A common construction is to represent each voxel as a top-dimensional $3$-cube and include all of its faces in the complex. A 3D CT scan comes with a natural filtration within the grayscale values of its voxels. A function is obtained on the resulting cubical complex $C$ as follows by extending the values of voxels to all the cubes $\omega \in C$:
\[
\eta'(\omega) := \min_{\omega \text{ face of $\tau$}} \eta(\tau).
\]
Given the cubical complex $C$ constructed from $I$ and let 
\[
C_{i} := \{\omega \in C \mid \eta'(\omega)\leq i\},
\]
be the $i$-th sublevel set of $C$. This yields a filtration of the cubical complex defined as $\{C_{i} \}_{i\in Im(I)}$ and indexed by the values of the function $\eta$. \cite{garin2019topological}.

\begin{figure}[ht]
\begin{center}
\includegraphics[width=1\linewidth]{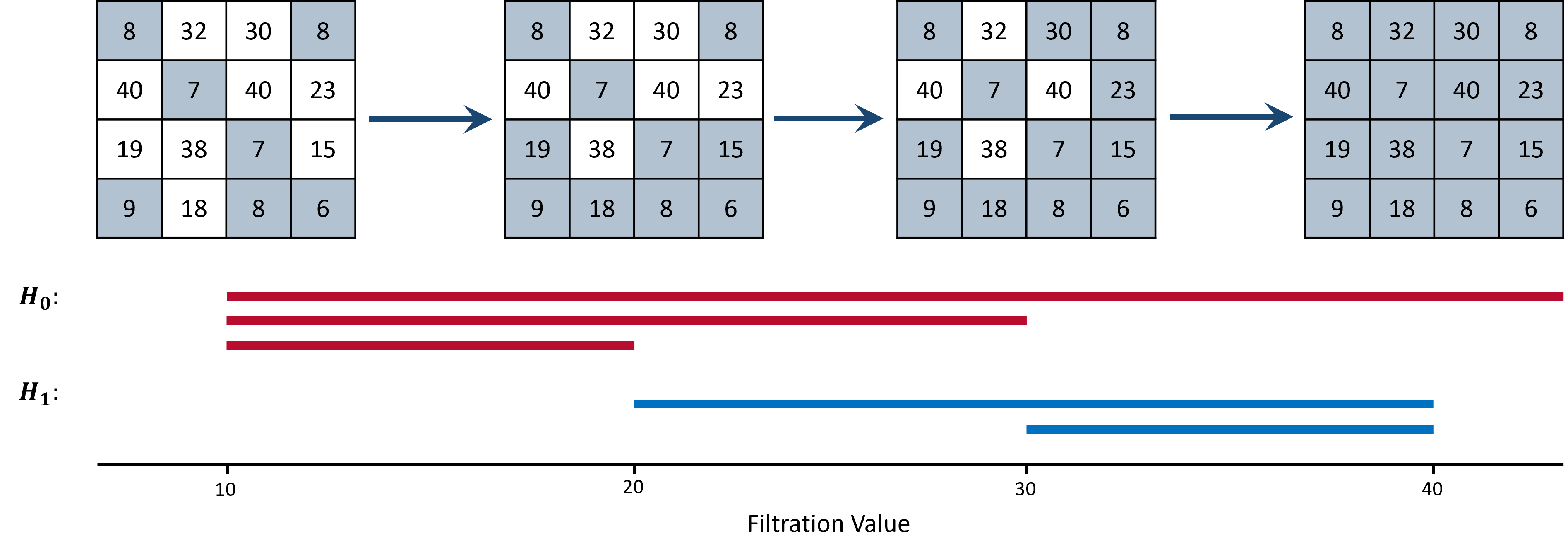}
\end{center}
\caption
{ \label{fig:cubical_complex_filtration} 
Cubical complex filtration and PBs representations in dimension zero and one for a 4x4 grayscale image patch. The red bars highlight the life line of the connected pixels, while the blue bars correspond to the loops appear between the connected pixels.}
\end{figure}

Since visualizing this filtration process on a 3D image is challenging on paper, for simplicity, this process is illustrated on a 4x4 patch of a grayscale image in Figure \ref{fig:cubical_complex_filtration}. Pixels with higher intensity are filtered and added to the complex as the filtration value is increased. Concurrently, the construction of PBs is appeared in the same figure, below the grid filtration process. The appearance (birth) and vanishing (death) of connected components and loops are highlighted by $H_0$ and $H_1$ respectively. In addition to these structures, the birth and death of cavities (voids), denoted by $H_2$, can also be computed when this procedure is applied on 3D CT images.

\subsection{Persistence Barcodes}
A persistence barcode can be expressed as
\[
B = [(p_1,q_1),(p_2,q_2),\cdots,(p_n,q_n)],
\]
where $B(p,q)$ corresponds to a bar with birth at $p$ and death at $q$. Note that a barcode has a multiset structure, meaning that an element $(p, q)$ in the barcode $B$ may occur multiple times \cite{hofer2019learning}. Therefore, the barcode $B$ has a multiplicity function $\mu: B \rightarrow \mathbb{Z}_{>0}$.

\begin{figure}[h]
\begin{center}
\includegraphics[width=1\linewidth]{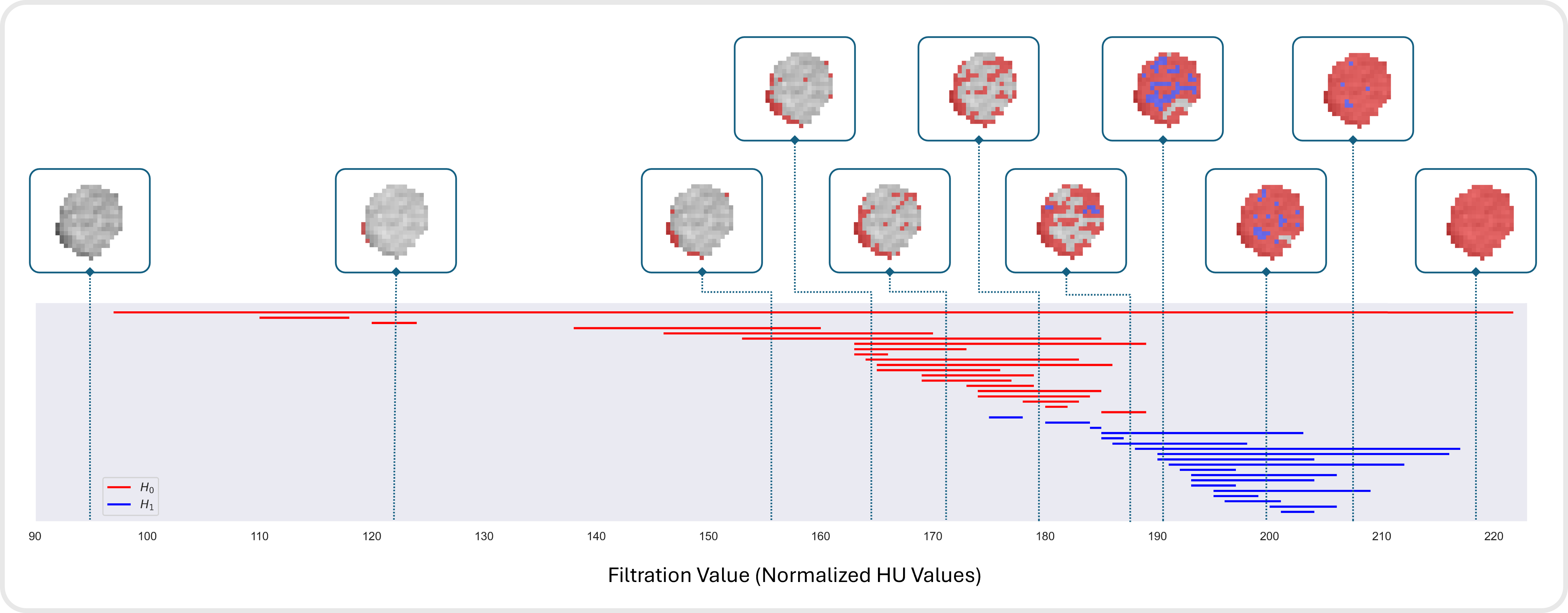}
\end{center}
\caption
{ \label{fig:tumour_cubical_conplex_filtration} 
A summary illustration of the process of cubical complex filtration on a sample 2D image, the ROI of a slice of a pancreatic tumour. For visualization purpose, we used the normalized HU values in this example. As the filtration value increases, the structure of connected pixels changes, and loop structures surrounded by the connected pixels appear and eventually vanish. The life timeline of connected components and loops are highlighted with the red and blue bars respectively.}
\end{figure}

Here we demonstrate the cubical complex filtration on a sample 2D image of a pancreatic tumour ROI due to the limitation of visualizing the 3D cubical complex filtration on the 3D tumour ROI. A summary of this procedure is illustrated in Figure \ref{fig:tumour_cubical_conplex_filtration}. The process utilizes an ordered list of distinct pixel intensity values within the image as the filtration parameter. Taking one value from the list at a time and filling in those pixels in the image whose intensity values are less than or equal to the current threshold value, connected components starts to appear, it can be either a single filled in pixel or pixels that are connected to each other. Following this filtration process and by increasing filtration thresholds, some of these connected pixels may join others quickly while some may persist to stay independent for longer. For example, neighbouring pixels with less varying intensities will join each other quickly whiles a pixel or a group of adjacent pixels whose intensity values stands out from the surrounding ones will stay as a single connected component for longer time. This cluster of pixels could be a darker spot in the tumour ROI revealing specific pathological patterns. Fortunately, we can keep track of all these short and long living connected components using persistent homology barcodes by representing the lifeline of each structure as a bar with its starting point (birth) to be the threshold where it starts to appear and the end point (death) indicating the threshold where it joins another cluster, and hence a collection of such bars forms a barcode. Other structures, beyond the connected pixels, whose evolution can be tracked during the filtration process are the loops. A loop structure could be a single or a group of trapped non-filled pixels surrounded by connected pixels whose values are less than the filtration threshold. This could be a lighter spot in the image or a larger region indicating separation of tumour (or soon to be a tumour) with respect to its surrounding. Utilizing this filtration process, both local and global topological features present in the image can be captured. Note that, when the filtration process is applied to the 3D tumour ROI, instead of the 2D slice only, beside the connected components and cycles, the evolution of enclosed voids, another higher dimensional structure, can also be tracked to generate the corresponding void-representing barcode.

\subsection{Vectorization}
The PBs acquired via a filtration process (here alpha or cubical complex), are not directly compatible with ML algorithms since they are not easily amenable to statistical analysis; for instance, their Fr\'echet mean is not unique \cite{hofer2019learning}. To overcome this bottleneck, barcodes are often embedded into vector spaces (e.g., Hilbert or Banach spaces), where well-defined inner products enable standard ML operations. Several vectorization techniques have been proposed in recent years to transform PBs into suitable feature vectors \cite{chung2022persistence, chevyrev2018persistence, berry2020functional, chintakunta2015entropy}. Persistent statistical vectorization is a frequently used approach to featurize the space of PBs which provides a compact summary of PBs and has shown robustness and success in comparison with other vectorization approaches to analysis digital images \cite{ali2023survey}. This approach constructs a feature vector by computing different statistical quantities from a given PB including the mean, median, standard deviation, full range, inter-quartile range, and the 10th, 25th, 75th, and 90th percentiles of the birth $p$, the death $q$, the lifespans $q-p$ and the midpoints of the bar intervals $[p, q]$ in a PB. Entropy and bar counts are also included to further extend the feature set. The entropy of $\mu$, introduced in \cite{chintakunta2015entropy}, is a real number defined as follows:
\[
E_\mu := -\sum_{[p,q] \in B} \mu_{p,q} \cdot \left(\frac{q-p}{L_\mu}\right) \cdot \log \left(\frac{q-p}{L_\mu}\right),
\]
where $L_\mu$ denotes the weighted sum 
\[
L_\mu := \sum_{[p,q] \in B} \mu_{p,q} \cdot (q-p).
\]

\subsection{Radiomic Features}
Radiomic features are quantitative measures that characterize tumour heterogeneity through the computational analysis of medical images \cite{gillies2016radiomics, lambin2012radiomics}. In this study, the default set of 107 radiomic features was extracted using pyradiomics package \cite{pyradiomics} and employed for benchmarking experiments. This feature set includes first-order statistics, shape-based features, gray level co-occurrence matrix (GLCM), gray level run length matrix (GLRLM), gray level size zone matrix (GLSZM), gray level dependence matrix (GLDM), and neighbourhood gray tone difference matrix (NGTDM) features.

\subsection{Patch-based TDA: The Proposed Approach}
At its core, patch-based TDA is an approach to constructing PH from volumetric data (e.g., 3D CT images), primarily focusing on transforming the data into a point cloud structure from which various topological features such as connected components, loops, and voids can be extracted. The intuition behind this technique is to reduce the effect of noisy voxels, present in 3D medical images, through image patch summarization while keeping important information from the patch including the location of the patch in the image and its voxel intensity values statistical summary. This process starts by taking cubic patches of size $n\!\times\!n\!\times\!n$ voxels from the input 3D image. selecting a cubic patch of, for instance $3\!\times\!3\!\times\!3$ voxels, from a 3D image, then flattening and extending it with its approximate central voxel coordinates will results in a vector containing 30 values. A straight forward approach would be to consider this vector as a point in 30 dimensions. However, this approach may not be ideal since this vector may contain values from the noisy voxels and building PH for CT scan of large sizes is a tedious task and requires extensive computational resources. Furthermore, the dimension of the resulting flattened patch will increase cubically as the patch size grows. To overcome this issue, we propose a patch-to-point conversion method serving as the backbone of the patch-based TDA approach to transform 3D images to point clouds. This patch-to-point conversion mechanism encompasses two sub-tasks: one to convert the patch coordinates to a compact representation, called coordinates encoding, and the second sub-task, called intensity encoding, to summarize the values of the flattened patch into a smaller vector.

\begin{figure}[htb]
\begin{center}
\includegraphics[width=1\linewidth]{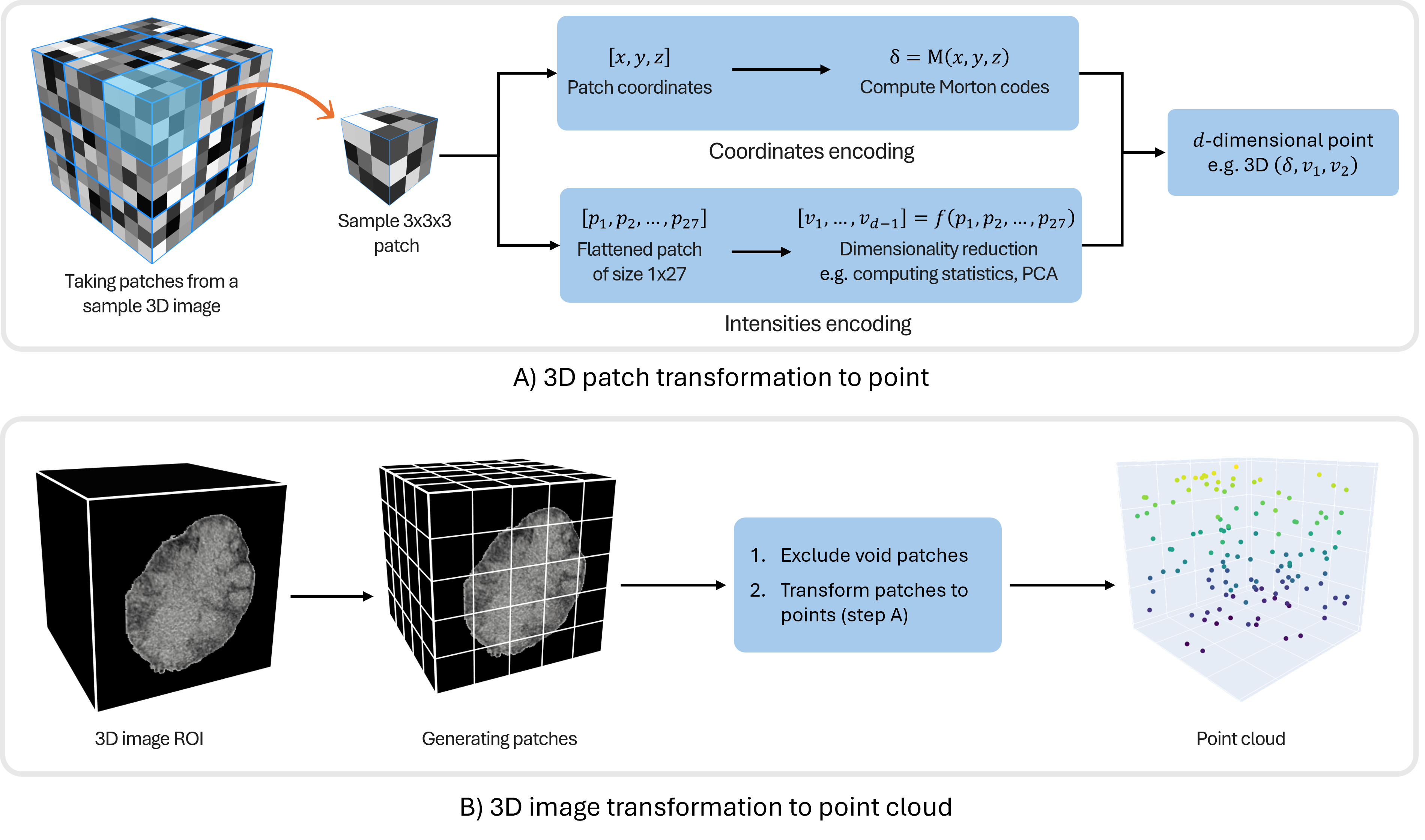}
\end{center}
\caption
{ \label{fig:point_cloud_pipeline} 
The process of generating a $d$-dimensional point from a 3D patch of $3\!\times\!3\!\times\!3$ voxels from a 3D image is demonstrated in part A, and the transformation of a 3D image ROI into a 3D point cloud is visualized in part B.
}
\end{figure}

For the coordinate encoding part, after preliminary experiments with a number of methods such as Morton code ($Z$-order curve) and $L_2$ norm, the Morton code algorithms \cite{morton1966computer} is utilized to compress the three coordinates ($x$, $y$ and $z$), corresponding to the approximate centre of the patch, to a single value. In parallel, for the second part of the pipeline, intensities encoding, two distinct approaches are considered: 1) employing PCA to reduce the dimension of the flattened patch and consequently utilizing $d$ PCA components to create the point, 2) computing statistical quantities (stats), for instance mean, mode and median, from the flattened patch to create the final point where each one of these stats corresponds to an axis in the space. This procedure is demonstrated in Figure \ref{fig:point_cloud_pipeline}, part A. Finally, applying the patch-to-point transformation techniques on all the patches taken from a volumetric ROI of a 3D image, a point cloud is obtained as depicted in Figure \ref{fig:point_cloud_pipeline}, part B. It is worth mentioning that, during this process only cubic patches with at least one voxel corresponding to the ROI are considered and converted to points contributing to the final point cloud, i.e, void or empty patches are discarded. This step is crucial to enhance the downstream computation of PH from the point cloud.

\begin{figure}[ht]
\begin{center}
\includegraphics[width=1\linewidth]{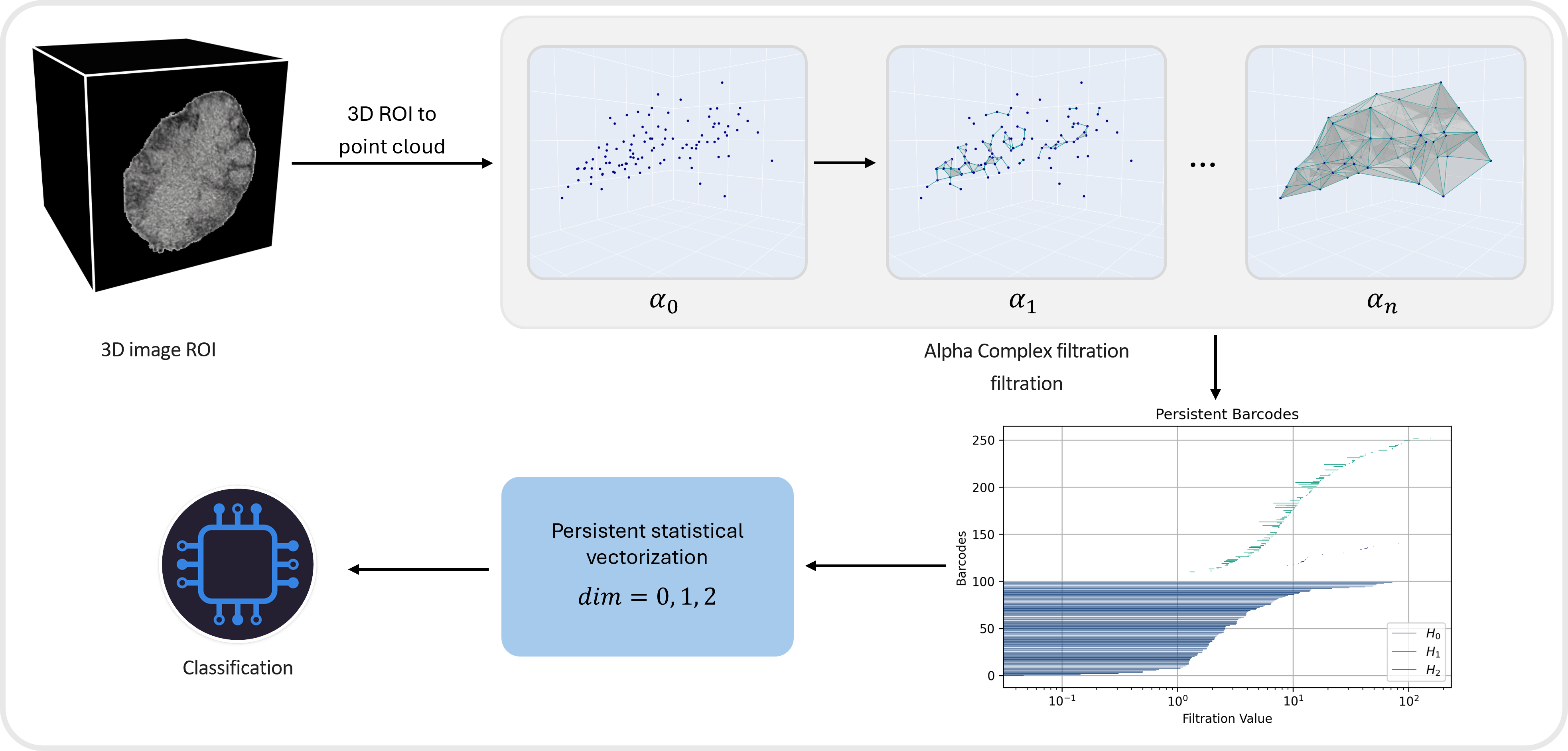}
\end{center}
\caption
{ \label{fig:patch_based_tda_pipeline} 
Patch-based TDA pipeline.
}
\end{figure}

The next step in the proposed approach is to compute PH from the resulting point clouds. In this study, alpha complex filtration \cite{edelsbrunner1983shape, boissonnat2018geometric} is utilized to compute PH in dimensions zero, one and two, corresponding to the number of connected components, loops, and cavities, respectively. Hence, obtaining three persistence barcodes. This follows by the vectorization process, computing persistent statistical features for each barcode, yielding three feature vectors. Finally, these feature vectors are concatenated and used as input to a ML classifier. This workflow is illustrated in Figure \ref{fig:patch_based_tda_pipeline}.

The first research question in our proposed patch-to-point method concerns determining the optimal patch size. A smaller patch size may include more noisy voxels, whereas a larger one could lead to the loss of important information. In this study, we define the patch size search space as $p=\{3, 4, 5, \dots, 10 \}$. The second research question investigates the optimal stats combination to summarize a flattened 3D patch (intensities encoding layer). To address this question, first a stats search space need to be defined to choose the combination from. We set this search space to $s=\{\text{mean, median, mode, standard deviation, interquartile range, entropy, range, minimum, maximum}\}$ which includes some of the most frequently used statistical quantities among others. In what follows, we discuss and show empirically what are the best $p$ and $s$ based on diverse 3D CT scan datasets described in the next section.

\subsection{Datasets}
\begin{figure} [ht]
\begin{center}
\includegraphics[width=1\linewidth]{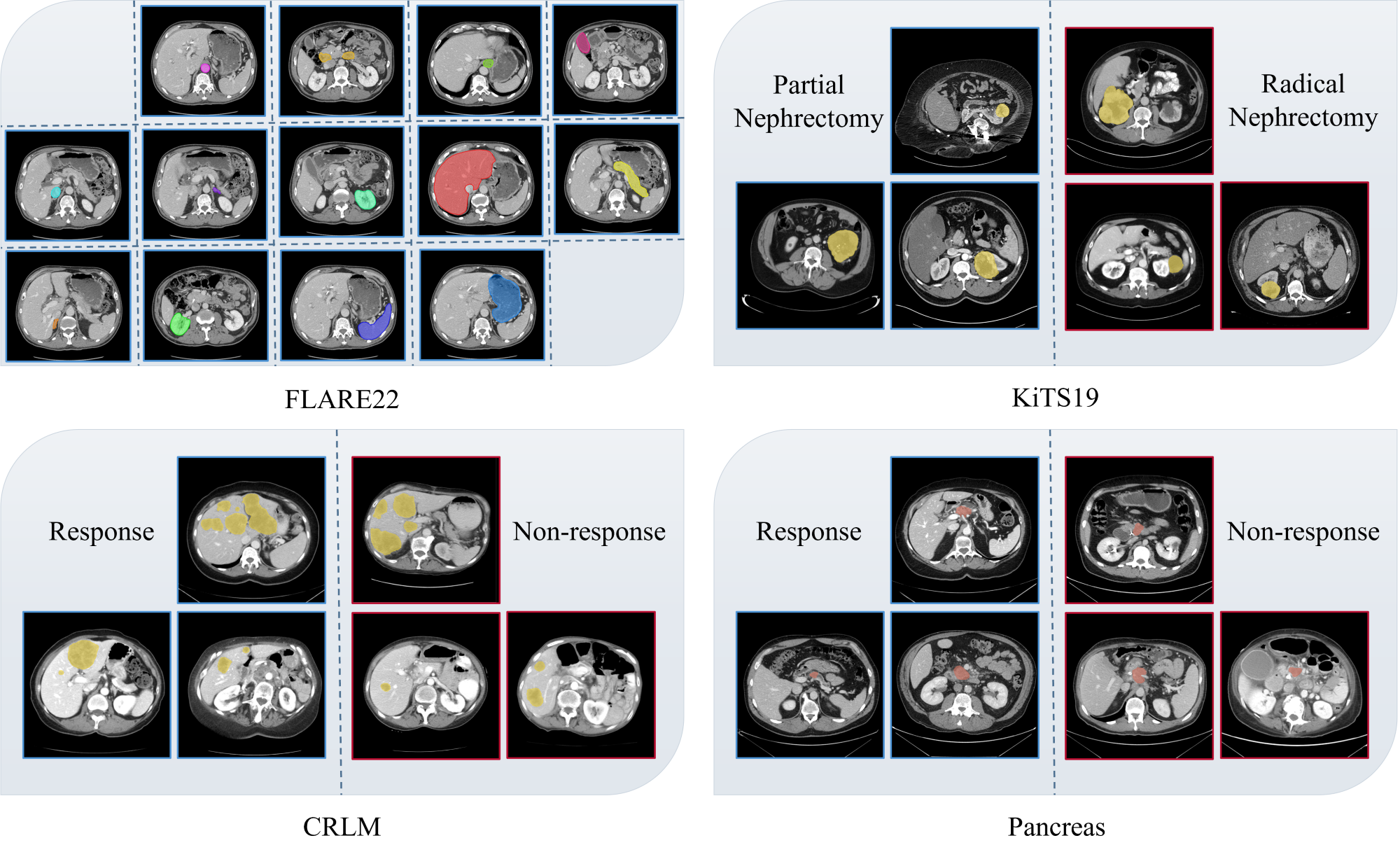}
\end{center}
\caption
{ \label{fig:dataset_sample} 
Sample slices of CT images from each dataset are visualized. Sample images of unique classes of each dataset are separated by the dashed lines.
}
\end{figure}

In this work, four CT datasets are utilized to assess the patch-based TDA approach and benchmark performance. These include the publicly available dataset of kidney tumours KiTS19 \cite{heller2019kits19}, the FLARE22 dataset from MICCAI FLARE22 challenge \cite{Ma2023-vo, ma2021abdomenct} which is a collection of abdominal CT images, and two in-house datasets. Sample CT slices of different classes of each dataset is illustrated in Figure \ref{fig:dataset_sample}. Table \ref{tab:datsets_summary} presents a summary of key details about each dataset.

\textbf{KiTS19}: this dataset was obtained from the GitHub repository of the kidney tumour segmentation challenge. It comprises abdominal CT scans and corresponding segmentation masks for 210 patients with kidney tumours. Among them, 70 underwent radical nephrectomy and 140 underwent partial nephrectomy. These surgical procedure types will serve as labels for downstream analysis.

\textbf{FLARE22}: the second dataset, FLARE22, is acquired from the MICCAI FLARE22 challenge. The dataset includes 50 abdominal CT scans alongside the segmentation masks of 13 different abdominal organs: liver, right kidney, spleen, pancreas, aorta, inferior vena cava, right adrenal gland, left adrenal gland, gallbladder, esophagus, stomach, duodenum and left kidney. With 50 CT scans and 13 different ROIs of organs, we obtained a total of 650 ROI images. Although classifying ROI of body organs may not be a clinically relevant task where segmentation masks are already provided, the dataset is solely utilized to design a multi-label classification task and assess the proposed patch-based approach and benchmark it against existing PH-construction methods.

\textbf{Colorectal liver metastases dataset (CRLM)}: this dataset includes 357 pre-treatment abdominal CT scans of patients diagnosed with unresectable CRLM and subsequently received chemotherapy at Memorial Sloan Kettering Cancer Centre (MSKCC). These patients were scanned again eight weeks after the treatment. An in-house developed segmentation model based on 3D U-Net architecture was utilized to obtain the segmentation masks of the tumours \cite{hamghalam2023attention}. Using pre and post treatment scans, response to therapy was measured by differences in the tumour size \cite{winter2018towards}. Therefore, patients were grouped as follows: 4 had complete response (CR), 152 had partial response (PR$\geq65\%$ tumour shrinkage), 24 had progressive disease (PD$\geq65\%$ tumour growth), and 177 had stable disease (SD, changes between PR and PD). For downstream analysis, CR and PR were grouped as ``response'' (156 patients), while SD and PD formed the ``non-response'' group (201 patients).

\textbf{Pancreas tumour dataset}: this final dataset is a collection of pre- and post-treatment abdominal CT scans from 97 patients who underwent neoadjuvant chemotherapy for pancreatic ductal adenocarcinoma (PDAC) at MSKCC. Eight patients were excluded from the analysis due to incomplete or missing pre-treatment scans, or missing RECIST labels. Pancreas tumours were manually segmented from the pre-treatment scans and verified by a board-certified radiologist. Using pre and post treatment scans, RECIST measurements were calculated by the radiologist. Based on these measurements, patients were grouped according to their response to therapy: 21 cases of PD, 43 cases of SD, and 25 cases of PR. There were no patients with a complete response. Furthermore, PD and SD were grouped into a single class, “non-response”. The final categorization of the dataset includes 25 “response” and 64 “non-response” cases.

\begin{table}[ht]
\caption{Summary of Key Details for Each CT Dataset Included in This Study.} 
\label{tab:datsets_summary}
\begin{center}  
\begin{tabular}{|l|l|l|l|}
\hline
Dataset & ROI & Number of Images & Unique Classes \& \# Samples/Class \\ \hline
KiTS19 & Kidney Tumours & 210 & \begin{tabular}[c]{@{}c@{}}Partial Nephrectomy (140), \\Radical Nephrectomy (70) \end{tabular} \\ \hline
FLARE22 & Abdomen Organs & 650 & 13 Organs (50 each)\\ \hline
CRLM & Tumours & 357 & Responder (156), Non-responder (201) \\ \hline
Pancreas Tumours & Tumours & 89 & Responder (25), Non-responder (64) \\ \hline
\end{tabular}
\end{center}
\end{table}

\subsection{Experiment Design and Implementation}
\subsubsection{Data Preprocessing}
The first pre-processing phase on each CT dataset was to account for the variable voxel sizes across CT scanners; we resampled all CT images to the average spacing within each dataset. The second phase involves applying the corresponding 3D segmentation masks to the CT images, acquired from the previous phase, masking out the targeted region of interest (ROI) and cropping the resulting volume with adequate padding on all sides. For the KiTS19 dataset, the final ROI, which consists of the kidney tumour only, is a 3D volume of $57\!\times\!220\!\times\!220$ voxels (i.e., 57 slices of $220\!\times\!220$ voxels each), whereas a 3D window of size $102\!\times\!290\!\times\!330$ voxels could accommodate the ROI of each organ in the preprocessed CT images from the FLARE22 dataset. For the CRLM and pancreas tumour datasets, the final ROIs were 3D volumes of $83\!\times\!321\!\times\!378$ and $33\!\times\!120\!\times\!120$ voxels respectively.

\subsubsection{Classification Pipeline Setup}
One of the most involved components in the classification pipeline of this study is the feature extraction phase, which includes the transformation of pre-processed ROI volume patches to point clouds, computation of PH in dimensions 0, 1, and 2 using alpha complex filtration, and featurization of PBs via persistent statistical vectorization followed by their concatenation across PH dimensions to obtain a single feature vector. To perform a fair assessment of the features obtained using the patch-based TDA approach, we employed a 5-fold cross-validation strategy for all classification tasks in this work. Stratified 5-fold splitting was used to ensure balanced class distribution across folds. In each fold, the data was standardized using z-score before being passed to a classifier. Five distinct classifiers, Support Vector Machine (SVM), Random Forest (RF), and K-Nearest Neighbours (KNN), Logistic Regression (LR) and Extreme Gradient Boosting (XGBoost) were employed.

\subsubsection{Patches and Stats Experiments}
The first conducted experiment in this study aimed to find the optimal combinations of patch size and stats. With the predefined search spaces of patch sizes and stats (comprising combinations of two or three stats), we performed a grid search, resulting in 960 experiments. In each trial, a patch size and stats combination were selected from the search spaces, the input 3D ROI volume was transformed into a point cloud and the rest of the classification pipeline was executed. This grid search experiment was consistently applied to each individual dataset. A systematic approach was employed to handle this grid search efficiently and finding the overall best combination of the patch size and stats. In the first stage of the grid search, at each trial, a nested 5-fold cross-validation (CV) was conducted with SVM, RF, KNN and LR only with a very shallow hyperparameters tuning (with inner CV). In the second stage, only the top 5\% of the combinations, with higher average performance across all the classifiers, were selected for the next phase. Finally, we experimented with the selected combinations using all the classifiers (including XGBoost), executing nested CV with deeper hyperparameters tuning.
 
\subsubsection{Patches and PCA Experiments}
The second experiment identified the optimal patch size when PCA was used to transform patches of the ROI volume into a point cloud. A grid search was conducted over the predefined patch size search space while employing PCA as the patch intensity summarization method. We restricted the dimensionality of the points in the point cloud generated via PCA to four dimensions (i.e., the Morton value and three PCA components). Increasing the dimension of the points beyond four will significantly increase the computational complexity of constructing PH. Therefore, in this experiment the dimension of the points were fixed to four. Since, this grid search only contains eight trials (patch size search space), we have utilized similar classification pipeline as in the final stage of the patch-stats grid search experiment.

\subsubsection{Benchmarking Experiments}
Finally, we benchmarked the patch-based PH construction method against the cubical complex, the classical approach of computing PH for grid-structured data as well as radiomic features. Two separate experiments were conducted: the first compared the methods in terms of classification performance, and the second assessed the computational efficiency of PH-based methods. Considering the volumetric nature of our data, unlike 2D images, the 3D version of cubical complex filtration was employed to construct PH in dimensions zero, one, and two, corresponding to connected voxels, loops, and voids between the voxels respectively. These PBs were then featurized using persistent statistical vectorization and concatenated. The resulting final feature vectors were subsequently passed into the same classification pipeline used in the previous experiments. To benchmark patch-based TDA against the cubical complex in terms of the time required to compute PBs, a sample 3D ROI of the preprocessed CT scans was selected from each dataset, and the average computation time was calculated over 100 trials. This experiment was conducted on a machine equipped with 64 GB of RAM and an Intel Core i7 processor, and there was no resource constraint for the calculations of PH performed by each method.

\section{Results}
In this study, we report relevant evaluation metrics including AUC, accuracy, sensitivity, specificity, and F1 score with standard deviation of each metric across the five outer folds of the nested CV experiments. The classification results of the grid search experiments on all four datasets are shown in Table \ref{tab:grid_search_results}. This includes the results of the grid search for optimal patch size and stats combination and the results of the second grid search for the best patch size with PCA. Results are sorted in terms of overall performance across all metrics and only the best results achieved with each method, stats and PCA, are reported for each dataset. It is apparent from the table that on all datasets, stats obtained better overall results than PCA. Notably, the PCA results are only marginally below the stats on FLARE22 dataset. It can be seen from the table that both methods of stats and PCA with FLARE22 and CRLM achieved better results with smaller patch sizes while for the rest of the reported results in the table larger patch sizes ($\ge 6$) were the winners. In terms of the optimal stats combination, both combinations of two and three stats can be observed in the results however there is no unique combination that would be the optimal choice of stats with all datasets. With regards to the choice of the classifier, LR appears as the clear winner.

\begin{table}[ht]
\centering
\begin{minipage}{0.48\textwidth}
  \centering
  \rotatebox{90}{%
    \begin{minipage}{\textheight}
      \centering
      \captionof{table}{Best Results for Each Dataset with Each Method of Stats and PCA for Patch Summarization.}
      \label{tab:grid_search_results}
      \vspace{1ex}
        \begin{tabular}{|c@{\hspace{0.12cm}}|c@{\hspace{0.12cm}}|@{\hspace{0.12cm}}c@{\hspace{0.12cm}}|@{\hspace{0.12cm}}c@{\hspace{0.12cm}}|@{\hspace{0.12cm}}c@{\hspace{0.12cm}}|@{\hspace{0.12cm}}c@{\hspace{0.12cm}}|@{\hspace{0.12cm}}c@{\hspace{0.12cm}}|@{\hspace{0.12cm}}c@{\hspace{0.12cm}}|@{\hspace{0.12cm}}c@{\hspace{0.12cm}}|@{\hspace{0.12cm}}c@{\hspace{0.12cm}}|}
            \hline
            
           Dataset & Method & Patch Size & Stats & Model & Accuracy ± Std. & AUC ± Std. & Sensitivity ± Std. & Specificity ± Std. & F1 ± Std. \\ \hline
           
            \multirow{2}{*}{KiTS19} & Stats & $6\!\times\!6\!\times\!6$ & mean, median, range & XGBoost & \textbf{87.1 ± 3.6} & \textbf{90.2 ± 4.8} & 74.3 ± 3.9 & \textbf{93.6 ± 4.7} & \textbf{79.5 ± 4.9} \\ \cline{2-10}
             & PCA & $9\!\times\!9\!\times\!9$ & - & LR & 83.8 ± 4.6 & 87.5 ± 4.5 & \textbf{82.9 ± 6.4} & 84.3 ± 7.4 & 77.5 ± 5.0 \\ \hline
             
            \multirow{2}{*}{FLARE22} & Stats & $4\!\times\!4\!\times\!4$ & mode, entropy & RF & \textbf{87.5 ± 1.5} & \textbf{99.2 ± 0.2} & \textbf{87.5 ± 1.5} & \textbf{99.0 ± 0.1} & \textbf{87.5 ± 1.4} \\ \cline{2-10}
             & PCA & $4\!\times\!4\!\times\!4$ & - & XGBoost & 86.2 ± 2.4 & 99.2 ± 0.3 & 86.2 ± 2.4 & 98.9 ± 0.2 & 86.1 ± 2.3 \\ \hline

            \multirow{2}{*}{\begin{tabular}[c]{@{}c@{}}Pancreas\\ Tumours\end{tabular}} & Stats & $6\!\times\!6\!\times\!6$ & iqr, entropy & LR & \textbf{75.2 ± 9.0} & \textbf{73.1 ± 9.3} & \textbf{60.0 ± 20.0} & \textbf{81.1 ± 10.6} & \textbf{57.2 ± 14.9} \\ \cline{2-10}
             & PCA & $10\!\times\!10\!\times\!10$ & - & LR & 69.6 ± 3.7 & 70.4 ± 9.7 & 60.0 ± 31.6 & 73.3 ± 12.0 & 49.4 ± 15.6 \\ \hline
             
            \multirow{2}{*}{CRLM} & Stats & $3\!\times\!3\!\times\!3$ & mean, entropy, max & XGBoost & \textbf{60.0 ± 4.2} & \textbf{60.9 ± 3.3} & \textbf{55.1 ± 5.5} & \textbf{63.7 ± 10.5} & \textbf{54.6 ± 2.6} \\ \cline{2-10}
             & PCA & $3\!\times\!3\!\times\!3$ & - & LR & 52.7 ± 2.9 & 50.8 ± 6.7 & 52.5 ± 10.4 & 52.7 ± 8.2 & 48.9 ± 5.9 \\ \hline
             
        \end{tabular}
    \end{minipage}
  }
\end{minipage}
\hfill
\begin{minipage}{0.48\textwidth}
  \centering
  \rotatebox{90}{%
    \begin{minipage}{\textheight}
      \centering
      \captionof{table}{Benchmarking Patch-Based TDA Against Cubical Complex and Radiomics}
      \label{tab:benchmarking}
      \vspace{1ex}
        \begin{tabular}{|c|c|c|c|c|c|c|c|}
        \hline
        
        Dataset & Method \& Parameters & Model & Accuracy ± Std. & AUC ± Std. & Sensitivity ± Std. & Specificity ± Std. & F1 ± Std. \\ \hline
        
        \multirow{4}{*}{KiTS19} & Radiomics & LR & 81.9 ± 5.2 & 89.2 ± 4.4 & \textbf{80.0 ± 3.2} & 82.9 ± 7.3 & 74.9 ± 5.7 \\ \cline{2-8}
        & Cubical Complex & XGBoost & 83.3 ± 6.1 & 87.7 ± 7.4 & 71.4 ± 20.8 & 89.3 ± 5.7 & 72.9 ± 12.3 \\ \cline{2-8} 
         & \begin{tabular}[c]{@{}c@{}}Patch-based TDA\\ ($6\!\times\!6\!\times\!6$, [mean, median, range])\end{tabular} & XGBoost & \textbf{87.1 ± 3.6} & \textbf{90.2 ± 4.8} & 74.3 ± 3.9 & \textbf{93.6 ± 4.7} & \textbf{79.5 ± 4.9} \\ \hline
        
        \multirow{4}{*}{FLARE22} & Radiomics & LR & 86.6 ± 2.3 & 99.1 ± 0.4 & 86.6 ± 2.3 & 98.9 ± 0.2 & 86.6 ± 2.3 \\ \cline{2-8}
        & Cubical Complex & LR & 77.1 ± 2.4 & 97.5 ± 0.8 & 77.1 ± 2.4 & 98.1 ± 0.2 & 76.9 ± 2.1 \\ \cline{2-8}
         & \begin{tabular}[c]{@{}c@{}}Patch-based TDA\\ ($4\!\times\!4\!\times\!4$, [mode, entropy])\end{tabular} & RF & \textbf{87.5 ± 1.5} & \textbf{99.2 ± 0.2} & \textbf{87.5 ± 1.5} & \textbf{99.0 ± 0.1} & \textbf{87.5 ± 1.4} \\ \hline

        \multirow{4}{*}{\begin{tabular}[c]{@{}c@{}}Pancreas\\ Tumours\end{tabular}} & Radiomics & SVM & 60.7 ± 4.1 & 63.2 ± 11.0 & 56.0 ± 16.7 & 62.6 ± 9.7 & 43.8 ± 7.0 \\ \cline{2-8}
        & Cubical Complex & LR & 55.0 ± 11.8 & 58.4 ± 9.9 & 68.0 ± 17.9 & 49.9 ± 18.9 & 45.9 ± 8.5 \\ \cline{2-8} 
         & \begin{tabular}[c]{@{}c@{}}Patch-based TDA\\ ($6\!\times\!6\!\times\!6$, [iqr, entropy])\end{tabular} & LR & \textbf{75.2 ± 9.0} & \textbf{73.1 ± 9.3} & \textbf{60.0 ± 20.0} & \textbf{81.1 ± 10.6} & \textbf{57.2 ± 14.9} \\ \hline

        \multirow{4}{*}{CRLM} & Radiomics & LR & 57.4 ± 5.0 & 60.3 ± 8.2 & 55.7 ± 10.7 & 58.7 ± 3.5 & 53.0 ± 7.9 \\ \cline{2-8}
        & Cubical Complex & SVM & 59.9 ± 6.3 & 59.7 ± 10.0 & \textbf{60.9 ± 7.5} & 59.1 ± 11.2 & \textbf{57.0 ± 5.6} \\ \cline{2-8}
         & \begin{tabular}[c]{@{}c@{}}Patch-based TDA\\ ($3\!\times\!3\!\times\!3$, [mean, entropy, max])\end{tabular} & XGBoost & \textbf{60.0 ± 4.2} & \textbf{60.9 ± 3.3} & 55.1 ± 5.5 & \textbf{63.7 ± 10.5} & 54.6 ± 2.6 \\ \hline \hline
         
        \multicolumn{3}{|c|}{Average Improvement} & 7.2 & 3.6 & 2.7 & 8.0 & 7.2 \\ \hline
        \end{tabular}
    \end{minipage}
  }
\end{minipage}
\end{table}

The results of benchmarking the patch-based TDA approach against the cubical complex method and radiomic features are shown in Table \ref{tab:benchmarking}. Overall, the patch-based approach outperformed both methods on all datasets. It is worth mentioning that the standard deviation of different metrics across the five folds experiments on all datasets is generally lower with the patch-based TDA than with the cubical complex approach which highlights the stability of the proposed approach.

Finally, we report the benchmarking results of the proposed approach against the cubical complex method in terms of PH computational efficiency. These results are depicted in Table \ref{tab:time_benchmarking}, where the average required time to compute PH over 100 trials is calculated for a sample 3D ROI from each dataset. It is apparent from the table that the patch-based TDA approach is significantly more efficient than the cubical complex method across all datasets especially with larger patch sizes which lead to smaller point clouds. For example, on the KiTS19 and pancreas tumours datasets, the proposed method is approximately 128 and 73 times faster than the cubical complex approach, respectively. Experimental observations indicate that with the patch-based approach, the PH computation time of 4D point clouds via the alpha complex is noticeably slower than its 3D counterpart, which can consequently affect the overall computation time of the proposed approach. Overall, the patch-based TDA approach provides improved results over cubical complex, in terms of both classification performance and time efficiency, particularly with high resolution ROI of CT images.

\clearpage
\begin{table}[ht]
\caption{The Results of Benchmarking the Patch-Based TDA Against the Cubical Complex Approach in Terms of the Required Time To Compute PH in Dimensions Zero, One and Two Across All Datasets} 
\label{tab:time_benchmarking}
\begin{center}
\begin{tabular}{|c|c|c|c|c|c|c|}
\hline
Dataset & Image Size & Method & Patch size & Stats & \begin{tabular}[c]{@{}c@{}}Points\\ Dimension\end{tabular} & \begin{tabular}[c]{@{}c@{}}Time\\ (s)\end{tabular} \\ \hline

\multirow{2}{*}{\begin{tabular}[c]{@{}c@{}}KiTS19\end{tabular}} & \multirow{2}{*}{$57\!\times\!220\!\times\!220$} & Patch-Based
& $6\!\times\!6\!\times\!6$ & mean, median, range & 4D & 0.3
\\ \cline{3-7} 
 &  & Cubical Complex & - & - & - & 33.4
\\ \hline

\multirow{2}{*}{FLARE22} & \multirow{2}{*}{$102\!\times\!290\!\times\!330$} & Patch-Based
& $4\!\times\!4\!\times\!4$ & mode, entropy & 3D & 2.5
\\ \cline{3-7} 
 &  & Cubical Complex & - & - & - & 124.0
\\ \hline

\multirow{2}{*}{\begin{tabular}[c]{@{}c@{}}Pancreas\\ Tumours\end{tabular}} & \multirow{2}{*}{$33\!\times\!120\!\times\!120$} & Patch-Based
& $6\!\times\!6\!\times\!6$ & iqr, entropy & 3D & 0.1
\\ \cline{3-7} 
 &  & Cubical Complex & - & - & - & 5.1
\\ \hline

\multirow{2}{*}{\begin{tabular}[c]{@{}c@{}}CRLM\end{tabular}} & \multirow{2}{*}{$83\!\times\!321\!\times\!378$} & Patch-Based
& $3\!\times\!3\!\times\!3$ & mean, entropy, max & 4D & 4.7
\\ \cline{3-7} 
 &  & Cubical Complex & - & - & - & 128.8
\\ \hline

\end{tabular}
\end{center}
\end{table}

\section{Python Package}
Code reusability is a crucial aspect in the computational implementation of a newly developed algorithm. Therefore, we provide a convenient python package, called Patch-TDA, to facilitate experimentation with the proposed approach. The package can be downloaded directly from the corresponding GitHub repository.
\footnote{https://github.com/dashtiali/patch-tda}

\section{Conclusion}
This work introduced a novel patch-based PH construction method for medical imaging data, a significant improvement over the cubical complex filtration, the traditional approach of building PH from grid-structured data and to overcome the computational burden of the latter method, in particular, with high resolution volumetric data. The process of transforming a 3D image patch to a point, a crucial component of the proposed method, was systematically analyzed and numerous experiments were conducted in this regard. Four datasets of 3D CT images comprehensively assessed the proposed approach. Our result suggests that utilizing stats as a patch-to-point transformation method, which eventually converts all patches of a 3D volume to a point cloud upon which PH can be constructed, yields better results in downstream analysis than using PCA. Furthermore, we benchmarked our method against the cubical complex filtration and radiomic features with respect to classification performance. It was concluded that, overall, the proposed technique outperformed the later two. Finally, a PH computational efficiency test was conducted to assess the two PH–based approaches. The proposed approach demonstrated significantly lower computational time compared to the cubical complex-based method to construct PH from 3D CT images.
A limitation of this work could be the lack of inclusion of further medical imaging datasets and modalities. There are several possible future research directions to further improve the performance of the patch-based TDA method, such as reducing the number of points in the point cloud via clustering methods. In addition, the method could also be deployed as a feature extraction mechanism, for instance, in an LSTM neural network to take advantage of the robustness of the method in capturing temporal information.

\section*{Acknowledgments}
This work was funded by National Institutes of Health grants R01CA233888 and SPORE grant P50CA257881.


\bibliographystyle{unsrt}  
\bibliography{references}

\end{document}